\documentclass{article}

\usepackage[final]{nips_2017}

\usepackage[utf8]{inputenc} 
\usepackage[T1]{fontenc}    
\usepackage{hyperref}       
\usepackage{url}            
\usepackage{booktabs}       
\usepackage{nicefrac}       
\usepackage{microtype}      
\usepackage[usenames, dvipsnames]{color}
\usepackage{bm}
\usepackage{graphicx}
\usepackage{dsfont}
\usepackage{comment}
\usepackage{amsmath,amssymb,amsxtra,amsfonts}
\usepackage{threeparttable}
\usepackage{algorithm}
\usepackage{algorithmic}
\usepackage{float}
\usepackage[caption = false]{subfig}

\newcommand{\Reals}{\mathbb{R}}

\newcommand{\binary}{\mathds{1}}
\DeclareMathOperator{\setintersection}{\bigcap}

\newcommand{\qed}{\hfill \ensuremath{\Box}}
\renewcommand{\Re}{\mathbb{R}}

\newcommand{\vecx}{\bm{x}}
\newcommand{\vecxi}{\bm{x}_{i}}
\newcommand{\yi}{y_{i}}
\newcommand{\Fnn}{F_{n-1}}

\newcommand{\wni}{w_{n,l}(\vecxi)}

\newcommand{\wzeroi}{w_{0}(\vecxi)}

\newcommand{\hn}{h_{n,l}}
\newcommand{\feat}{\mathcal{X}}
\newcommand{\capr}{\bm{\mathcal{R}}_{n,l}}

\newcommand{\rbn}{R_{n}}

\newcommand{\rcbn}{R^{c}_{n}}

\newenvironment{proof}[1][Proof]{\begin{trivlist}
\item[\hskip \labelsep {\bfseries #1}]}{\end{trivlist}}

\newtheorem{theorem}{Theorem}

\newtheorem{remark}{Remark}
\newtheorem{lemma}{Lemma}

\title{Tree-Structured Boosting: Connections Between Gradient Boosted Stumps and Full Decision Trees}

\author{
Jos\'e Marcio Luna $^1$ \hspace{0.5em}
Eric Eaton $^2$ \hspace{0.5em}
Lyle H. Ungar $^2$ \hspace{0.5em} Eric Diffenderfer
$^1$ \\
{\bf Shane T. Jensen $^3$ \hspace{0.5em} Efstathios D. Gennatas $^4$ \hspace{0.5em}  Mateo Wirth $^3$} \\
{\bf Charles B. Simone II  $^5$ \hspace{0.5em} Timothy D. Solberg $^4$ \hspace{0.5em}
Gilmer Valdes $^4$} \\[0.2em]
$^1$ Dept. of Radiation Oncology, University of Pennsylvania\\
  \texttt{\{Jose.Luna,Eric.Diffenderfer\}@uphs.upenn.edu}\\
$^2$ Dept. of Computer and Information Science, University of Pennsylvania\\
  \texttt{\{eeaton,ungar\}@cis.upenn.edu}\\
$^3$ Dept. of Statistics, University of Pennsylvania\\
  \texttt{\{stjensen,mwirth\}@wharton.upenn.edu} \\
$^4$ Dept. of Radiation Oncology, University of California, San Francisco\\
  \texttt{\{Efstathios.Gennatas,Timothy.Solberg,Gilmer.Valdes\}@ucsf.edu} \\
$^5$ Dept. of Radiation Oncology, University of Maryland Medical Center\\
  \texttt{CharlesSimone@umm.edu}
}
\begin{document}

\maketitle

\section{Introduction}

{\em Classification And Regression Tree} (CART) analysis \cite{breiman1984classification} is a well-established statistical learning technique, which has been adopted by numerous other fields for its model interpretability, scalability to large data sets, and connection to rule-based decision making \cite{Loh2014Fifty}.  CART builds a model by recursively partitioning  the instance space, labeling each partition with either a predicted category (in the case of classification) or real-value (in the case of regression).   Despite their widespread use, CART models often have lower predictive performance than other statistical learning models, such as kernel methods and ensemble techniques~\cite{Caruana2006Empirical}.
Among the latter, boosting methods were developed as a means to train an ensemble of weak learners (often CART models) iteratively into a high-performance predictive model, albeit with a loss of model interpretability.  In particular, {\em gradient boosting} methods~\cite{friedman2001greedy} focus on iteratively optimizing an  ensemble's prediction to increasingly match the labeled training data.   Historically these two categories of approaches, CART and gradient boosting, have been studied separately, connected primarily through CART models being used as the weak learners in boosting. This paper investigates a deeper and surprising connection between full interaction models like CART and additive models like gradient boosting, showing that the resulting models exist upon a spectrum.  In particular, this paper includes the following contributions:
\vspace{-1em}
\begin{itemize}
\item We introduce {\em tree-structured boosting} (TSB) as a new mechanism for creating a hierarchical ensemble model that recursively partitions the instance space, forming a perfect binary tree of weak learners.  Each path from the root node to a leaf represents the outcome of a gradient boosted stumps (GBS) ensemble for a particular partition of the instance space.
\item We prove that TSB generates a continuum of single tree models with accuracy between CART and GBS, controlled via a single tunable parameter. In effect, TSB bridges between CART and GBS, identifying never-before-seen connections between additive  and full interaction models.
\item This result is verified empirically, showing that this hybrid combination of CART and GBS can outperform either approach individually in terms of accuracy and/or interpretability while building a single tree.  Our experiments also provide  insight into the continuum of models revealed by TSB.
\end{itemize}


\section{Connections between CART and Boosting}
\label{sect:Background}

\begin{figure}[tb!]
\begin{center}
\includegraphics[width=0.7\textwidth, clip, trim = 0in 3.2in 0in 3.2in]{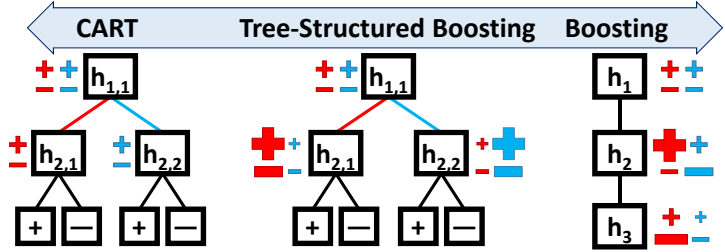}
\end{center}
\caption{CART, tree-structured boosting, and standard GBS, each given four training instances (blue and red points). The size of each point depicts its weight when used to train the adjacent node.}
\label{fig:TreeStructuredBoosting}
\end{figure}

Assume we are given a training set $(\bm{X},\bm{y}) = \{(\vecxi,y_i)\}_{i=1}^N$, where each $d$-dimensional $\vecxi \in \bm{X} \subseteq \mathcal{X}$ has a corresponding label $y_i \in \mathcal{Y}$, drawn i.i.d from a unknown distribution $\mathcal{D}$.  In a classification setting,  $\mathcal{Y} = \{\pm 1\}$; in regression, $\mathcal{Y} = \Reals$. The goal is to learn a function $F : \mathcal{X} \mapsto \mathcal{Y}$ that will perform well in predicting the label on new examples drawn from $\mathcal{D}$.  CART analysis recursively partitions $\mathcal{X}$, with $F$ assigning a single label in $\mathcal{Y}$ to each partition.  In this manner, there is full interaction between each component of the model. Different branches of the tree are trained with disjoint subsets of the data, as shown in Figure~\ref{fig:TreeStructuredBoosting}.

In contrast, boosting iteratively trains an ensemble of $T$ weak learners $\{h_t : \mathcal{X} \mapsto \mathcal{Y} \}_{t=1}^T$, such that the model\footnote{In classification, $F$ gives the sign of the prediction. CART models are often used as the weak learners.} is a weighted sum of the weak learners' predictions \mbox{$F(\vecx) = \sum_{t=1}^T \rho_t h_t(\vecx)$} with weights $\bm{\rho} \in \Reals^T$.  Each boosted weak learner is trained with a different weighting of the {\em entire} data set, unlike CART, repeatedly emphasizing mispredicted instances to induce diversity (Figure~\ref{fig:TreeStructuredBoosting}).  Gradient boosting with decision stumps or simple regression creates a pure additive model, since each new ensemble member serves to reduce the residual of previous members \cite{Friedman1998Additive,friedman2001greedy}. Interaction terms can be included in the overall ensemble by using more complex weak learners, such as deeper trees.

As shown by Valdes et al.~ \cite{valdes2016mediboost}, classifier ensembles with decision stumps as the weak learners, $h_t(\vecx)$, can be trivially rewritten as a complete binary tree of depth $T$, where the decision made at each internal node at depth $t\!-\!1$ is given by $h_t(\vecx)$, and the prediction at each leaf is given by $F(\vecx)$.  Intuitively, each path through the tree represents the same ensemble, but one that tracks the unique combination of predictions made by each member.

\subsection{Tree-Structured Boosting}
\label{sect:TreeStructuredBoosting}

 This interpretation of boosting lends itself to the creation of a tree-structured ensemble learner that bridges between CART and gradient boosting.  The idea behind tree-structured boosting~(TSB) is to grow the ensemble recursively, introducing diversity through the addition of different sub-ensembles after each new weak learner.   At each step, TSB first trains a weak learner on the current training set $\{(\vecxi,y_i)\}_{i=1}^N$ with instance weights $\bm{w} \in \Reals^N$, and then creates a new sub-ensemble for each of the weak learner's outputs. 
 Each sub-ensemble is subsequently trained on the full training set, but instances corresponding to the respective branch are more heavily weighted during training, yielding diverse sub-ensembles (Figure~\ref{fig:TreeStructuredBoosting}, middle). This process proceeds recursively until the depth limit is reached. Critically, this approach identifies clear connections between CART and Gradient Boosted Stumps (GBS): as the re-weighting ratio is varied, tree-structured boosting produces a spectrum of models with accuracy between CART and TBS at the two extremes.

 The complete TSB approach is detailed as Algorithm~\ref{alg:TSB}. The parameter $\lambda$ used in step 8 of the algorithm provides the connection between CART and GBS, \emph{i.e.,} TSB converges to CART as $\lambda \rightarrow 0$ and converges to GBS as $\lambda \rightarrow \infty$. 
 Theoretical analysis of TSB, given in the supplemental material, shows how TSB bridges between CART and GBS.

\begin{algorithm}[t!]
\caption[Tree-StructuredBoosting]
{
$\mbox{TreeStructuredBoosting}\!\left (\bm{X}, \bm{y}, \bm{w}, \lambda, n, T, R, F_{n-1} \right)$\label{alg:MediBoost}\\
{\bf Inputs:}\hspace{.7em} \parbox[t]{4.9in}{
training data $(\bm{X}, \bm{y}) = \{(\vecxi,y_i)\}_{i=1}^N$,
instance weights $\bm{w} \in \Reals^N$ (default: $w_i = \frac{1}{N}$),
$\lambda \in [0, +\infty]$,
node depth $n$ (default: $0$),
max height $T$,
node domain $\bm{\mathcal{R}}$ (default: $ \mathcal{X}$),
prediction function $F_{n-1}(\vecx)$  (default: $F_0(\vecx) = \bar{y}$)
}
{\bf Outputs:} the root node of a hierarchical ensemble}
\label{alg:TSB}

\begin{algorithmic}[1]

\STATE If $n > T$, return a prediction node $l_n$ that predicts the weighted average of $\bm{y}$ with weights $\bm{w}$

\STATE Create a new subtree root $l_n$ to hold a weak learner

\STATE Compute negative gradients $\left\{ \tilde{y}_i = - \frac{\partial \ell(y_i, F_{n-1}(\vecxi))}{\partial F_{n-1}(\vecxi)}\right\}_{i=1}^N$

\STATE Fit weak classifier $h_n'(\vecx):\mathcal{X}\mapsto \mathcal{Y}$ by solving
$
h_n' \leftarrow \arg\min_{h,\beta} \sum_{i=1}^N w_i (\tilde{y}_i- \beta h(\vecxi))^{2}
$, where $\beta$ is a scalar defining the additive expansion to estimate $F(\vecx)$.

\STATE Let $\{P_n, P_n^c\}$ be the partitions induced by $h_n'$.

\STATE
$
\rho_{nl} \leftarrow \arg\min_{\rho} \sum_{i=1}^N w_{i}  \left( y_i - F_{n-1}(\vecxi) - \rho h'_{nl}(\vecxi) \right)^2
$

\STATE Update the current function estimation
$
F_n(\vecx) = F_{n-1}(\vecx) + \rho h'_n(\vecx)
$

\STATE Update the left and right subtree instance weights, and normalize them:
{\vspace{-0.5em}\begin{align*}
    w_i^{(\mathrm{left})} &\propto w_i\Bigl(\lambda + \binary[\vecxi \in P_n]\Bigr) & w_i^{(\mathrm{right})} &\propto w_i\Bigl(\lambda + \binary[\vecxi \in P_n^c]\Bigr)
\end{align*}
\vspace{-1.25em}}

\STATE If $\bm{\mathcal{R}} \setintersection P_n \ne \emptyset $, compute the left subtree recursively:\\
~\hspace{1em}$l_n.\mathrm{left}\leftarrow\text{TreeStructuredBoosting}\!\left(\bm{X},\bm{y}, \lambda, \bm{w}^{(\mathrm{left})}, \bm{\mathcal{R}} \setintersection P_n, n+1, T, F_{n}  \right)$

\STATE If $\bm{\mathcal{R}} \setintersection P_n^c \ne \emptyset $, compute the right subtree recursively:\\
~\hspace{1em}$l_n.\mathrm{right}\leftarrow\text{TreeStructuredBoosting}\!\left(\bm{X},\bm{y}, \lambda, \bm{w}^{(\mathrm{right})}, \bm{\mathcal{R}} \setintersection P_n^c, n+1, T, F_{n}  \right)$

\end{algorithmic}
\end{algorithm}

\section{Experiments}
In a first experiment, we use real-world data to evaluate the classification error of TSB for different values of $\lambda$. We then examine the behavior of the instance weights as $\lambda$ varies in a second experiment.

\subsection{Assessment of TSB Model Performance versus CART and GBS}
In this experiment, we use four life science data sets from the UCI repository \cite{UCIMLRepository}: Breast Tissue, Indian Liver Patient Dataset (ILPD), SPECTF Heart Disease, and Wisconsin Breast Cancer.  These data sets are all binary classification tasks and contain only numeric attributes with no missing values. We measure the classification error as the value of $\lambda$
increases from $0$ to $\infty$. In particular, we assess 10 equidistant error points corresponding to the in-sample
and out-of-sample errors of the generated TSB trees, and plot the transient behavior of the classification errors
as functions of $\lambda$. The goal is to
illustrate the trajectory of the classification errors of TSB, which is expected to approximate the performance of CART as $\lambda \rightarrow 0$, and to converge asymptotically to
GBS as $\lambda \rightarrow \infty$.

To ensure fair comparison, we assessed the classification accuracy of CART and GBS for different depth and learning rate values over 5-fold cross-validation. As a result, we concluded that a tree/ensemble depth of 10 offered near-optimal accuracy, and so use it for all algorithms.  The binary classification was carried out using the negative binomial log-likelihood as the loss function, similar to LogitBoost \cite{friedman2001greedy}, which requires an additional learning rate (shrinkage) factor, via Algorithm 1.
\vspace{-1em}
\begin{table}[b!]
  \caption{Data Set Specifications}
  \label{dataset-table}
  \centering
  \begin{tabular}{lccc}
    \toprule
    Data Set     & \# Instances     & \# Attributes	&	TSB Learning Rate\\
    \midrule
    Breast Tissue	& 106	&	9	&	0.3\\
    ILPD		& 583	& 	9	&	0.3\\
    SPECTF		& 80	&	44	&	0.3\\
    Wisconsin Breast Cancer	& 569	& 	30	&	0.7\\
    Synthetic 		& 100	& 	2	&	0.1\\
    \bottomrule
  \end{tabular}
\end{table}

For each data set, the experimental results were averaged over 20 trials of 10-fold cross-validation over the data, using $90\%$ of the samples for training and the remaining $10\%$ for testing in each experiment.  The error bars in the plots denote the standard error at each sample point.

\begin{figure}
\subfloat[Breast Tissue]{\includegraphics[height = 1.1in]{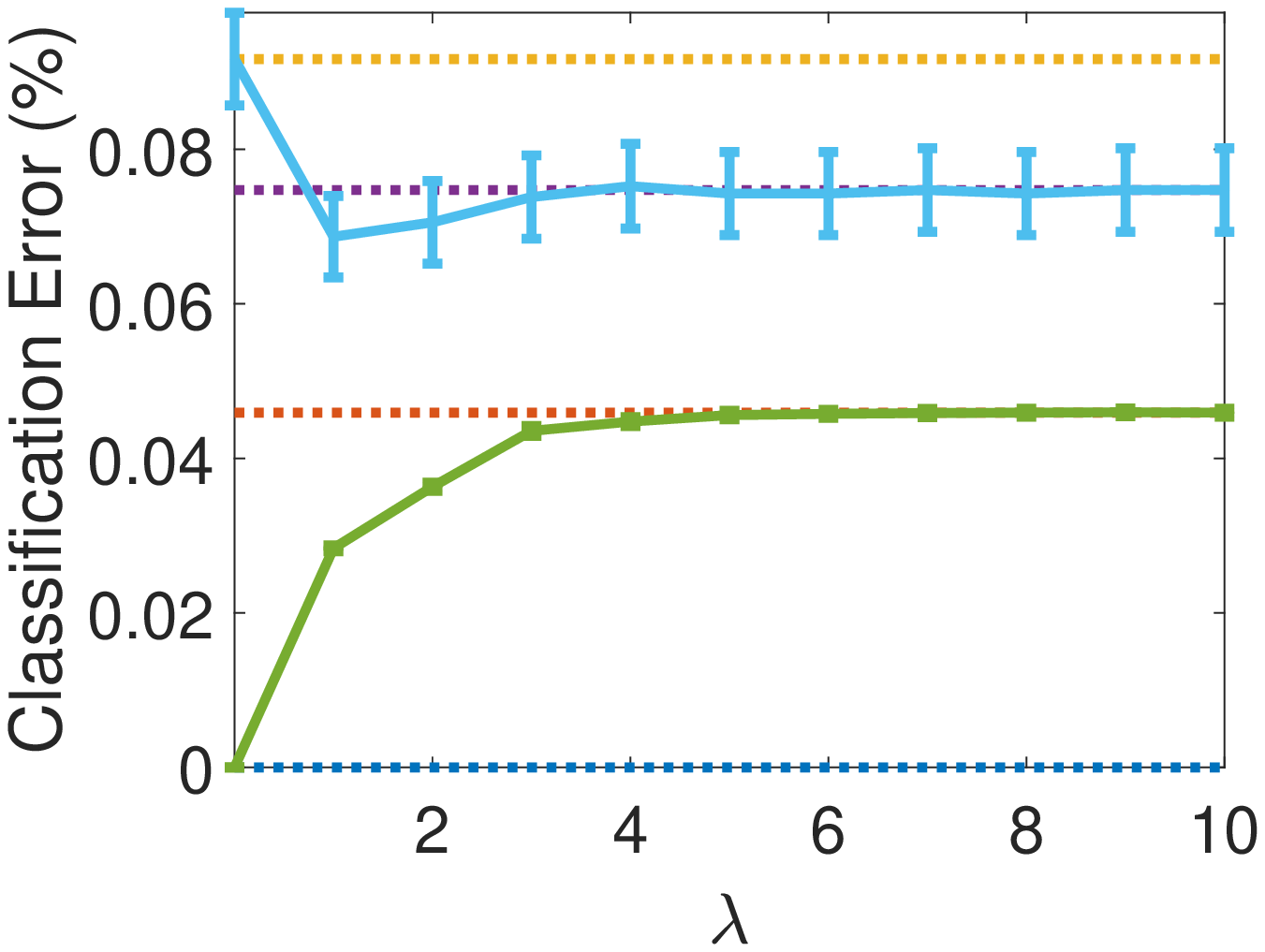}}
\subfloat[ILPD]{\includegraphics[height = 1.1in,clip,trim=.4in 0in 0in 0in]{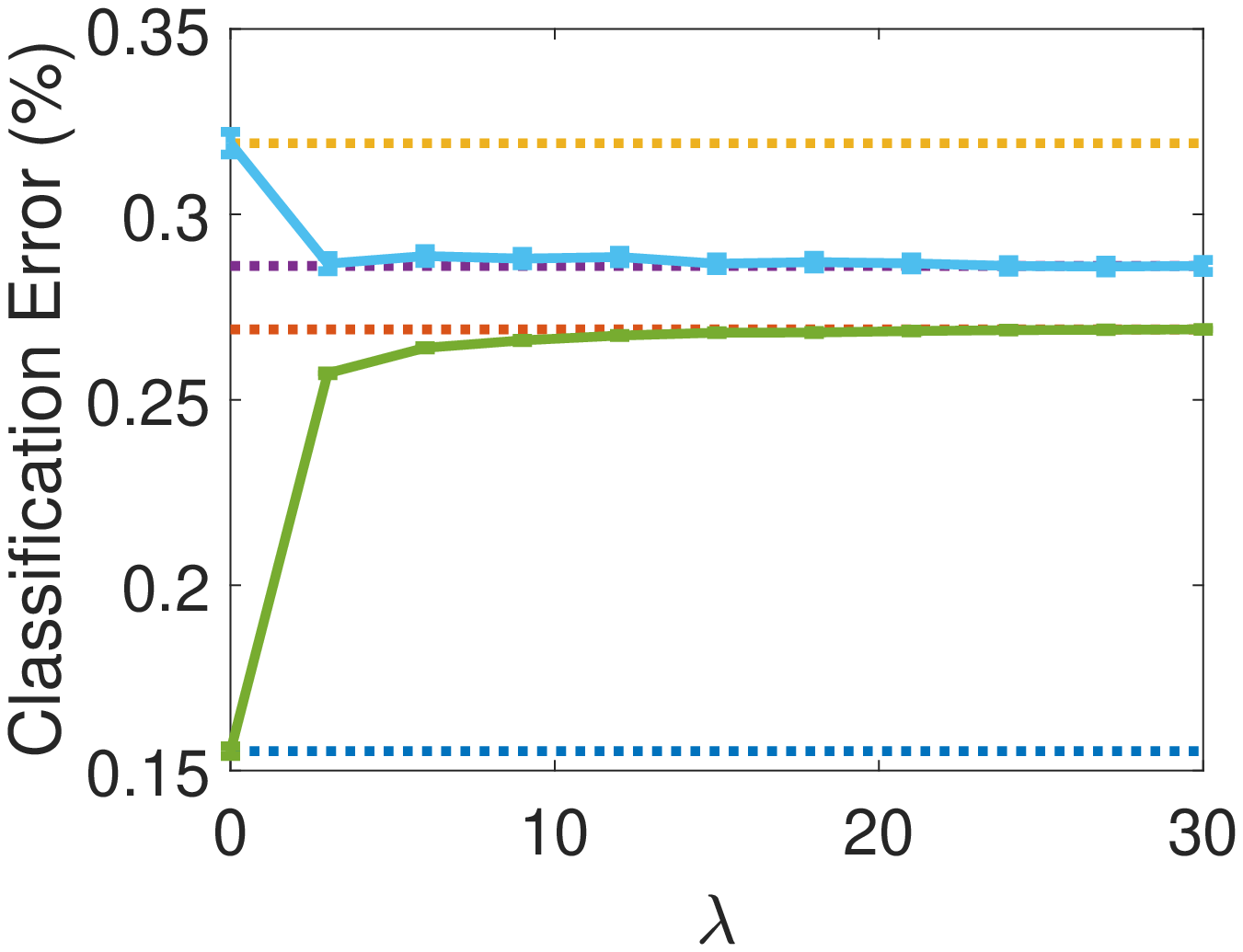}} 
\subfloat[SPECTF]{\includegraphics[height = 1.1in,clip,trim=.4in 0in 0in 0in]{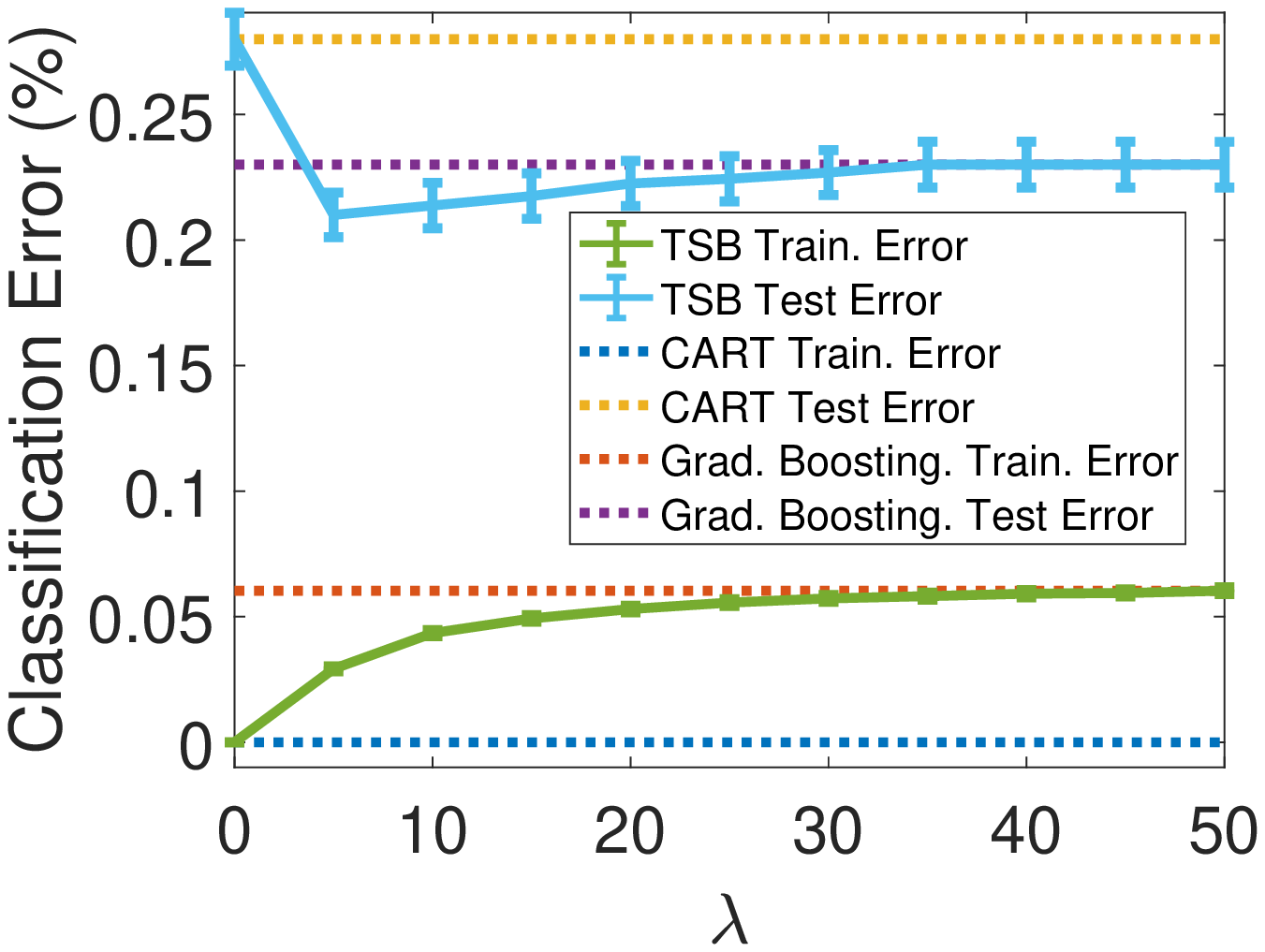}}
\subfloat[Breast Cancer]{\includegraphics[height = 1.1in,clip,trim=.4in 0in 0in 0in]{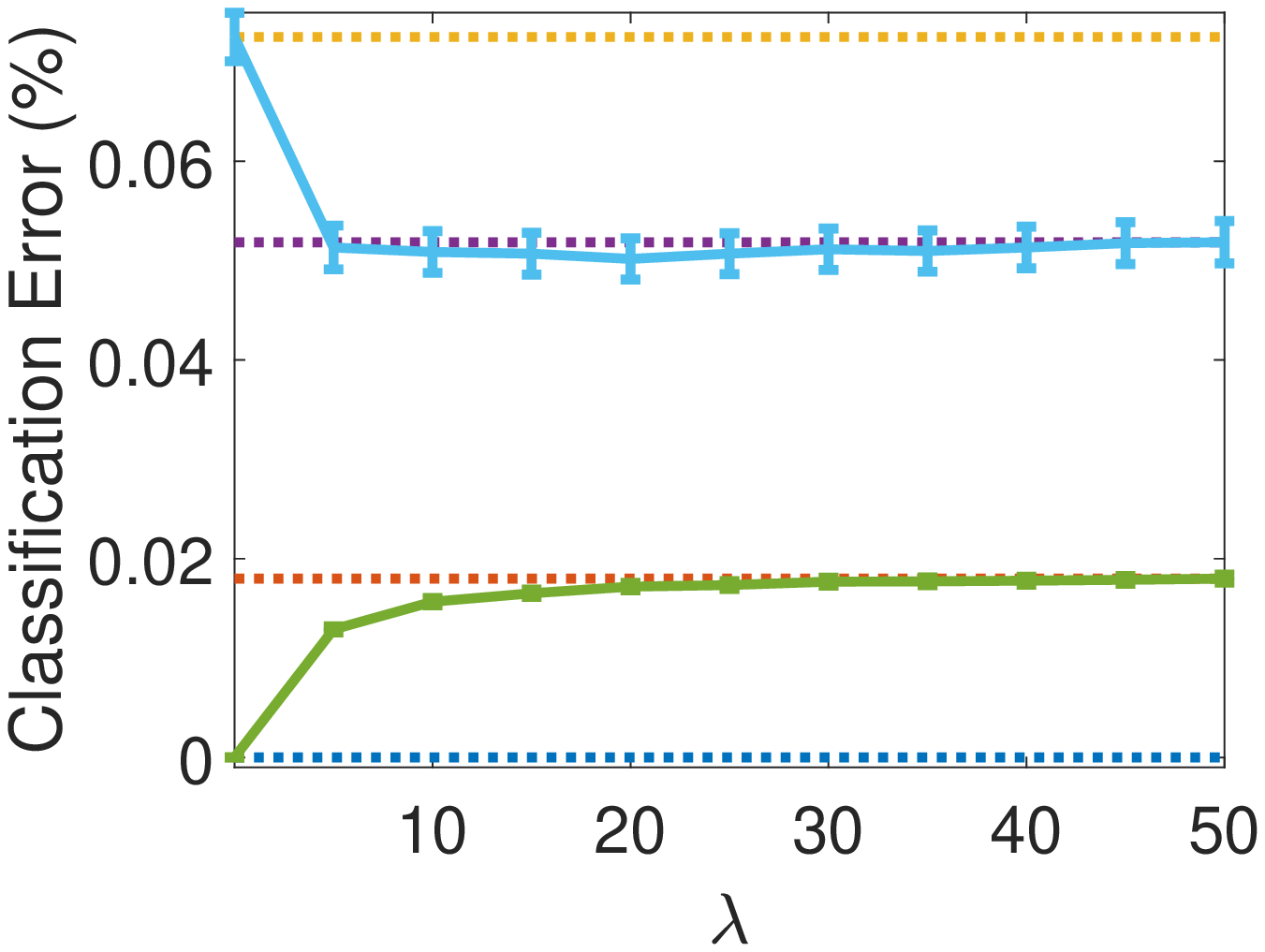}}
\caption{In-sample and out-of-sample classification errors for different values of $\lambda$. All plots share the same legend and vertical axis label. (Best viewed in color.)}
\label{error-plots}
\end{figure}

The results are presented in Figure~\ref{error-plots}, showing that the classification error of TSB approximates the CART and GBS errors in the limits of $\lambda$. As expected, increasing lambda generally reduces overfitting.  However, note that for each data set except ILPD, the lowest test error is achieved by a TSB model {\em between} the extremes of CART and GBS.  This reveals that hybrid TSB models can outperform either of CART or GBS alone.

\subsection{Effect of $\lambda$ on the Instance  Weights}
\begin{figure}
\subfloat[$\lambda = 0$]{\includegraphics[width = 1.36in]{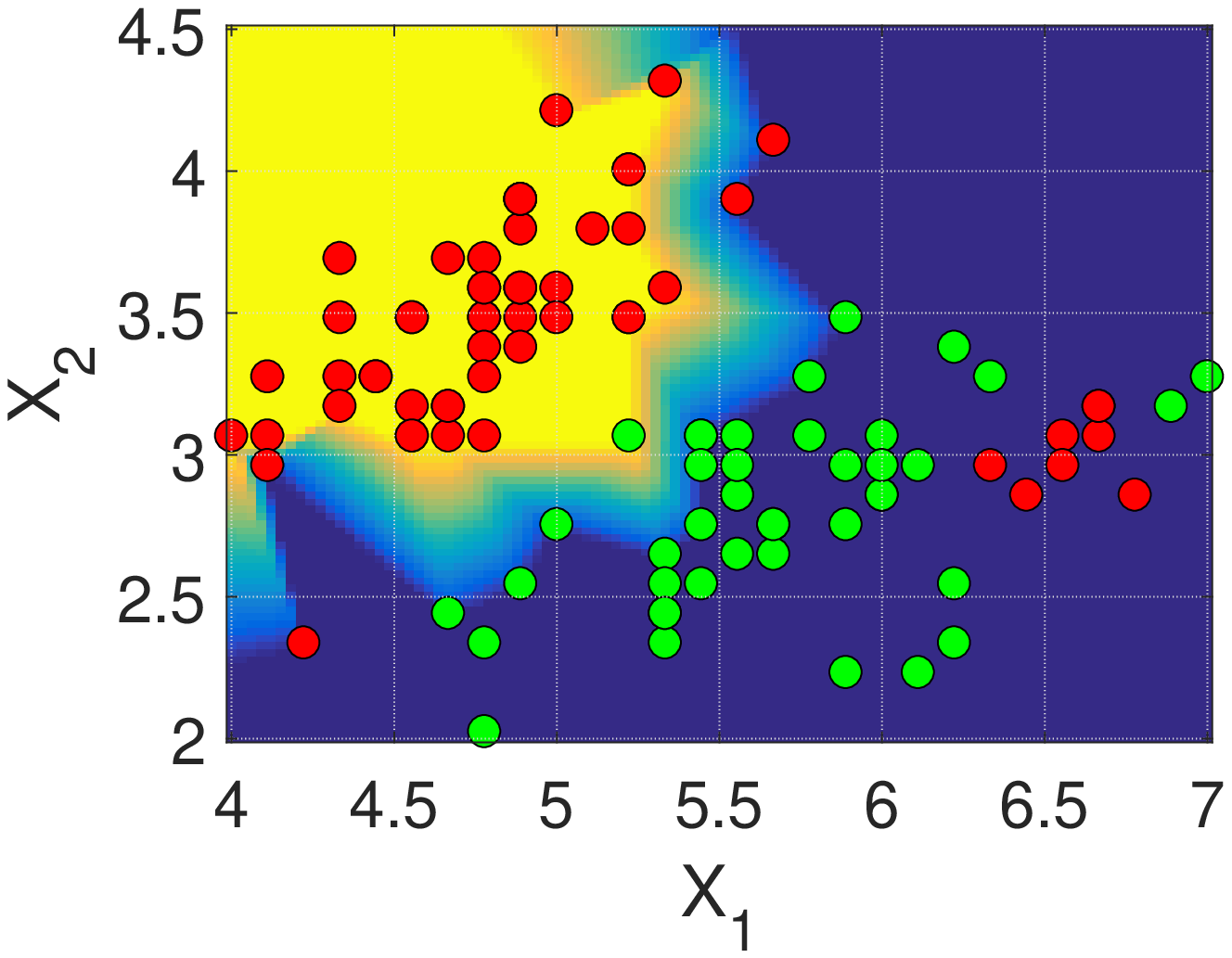}}
\subfloat[$\lambda = 0.2$]{\includegraphics[width = 1.36in]{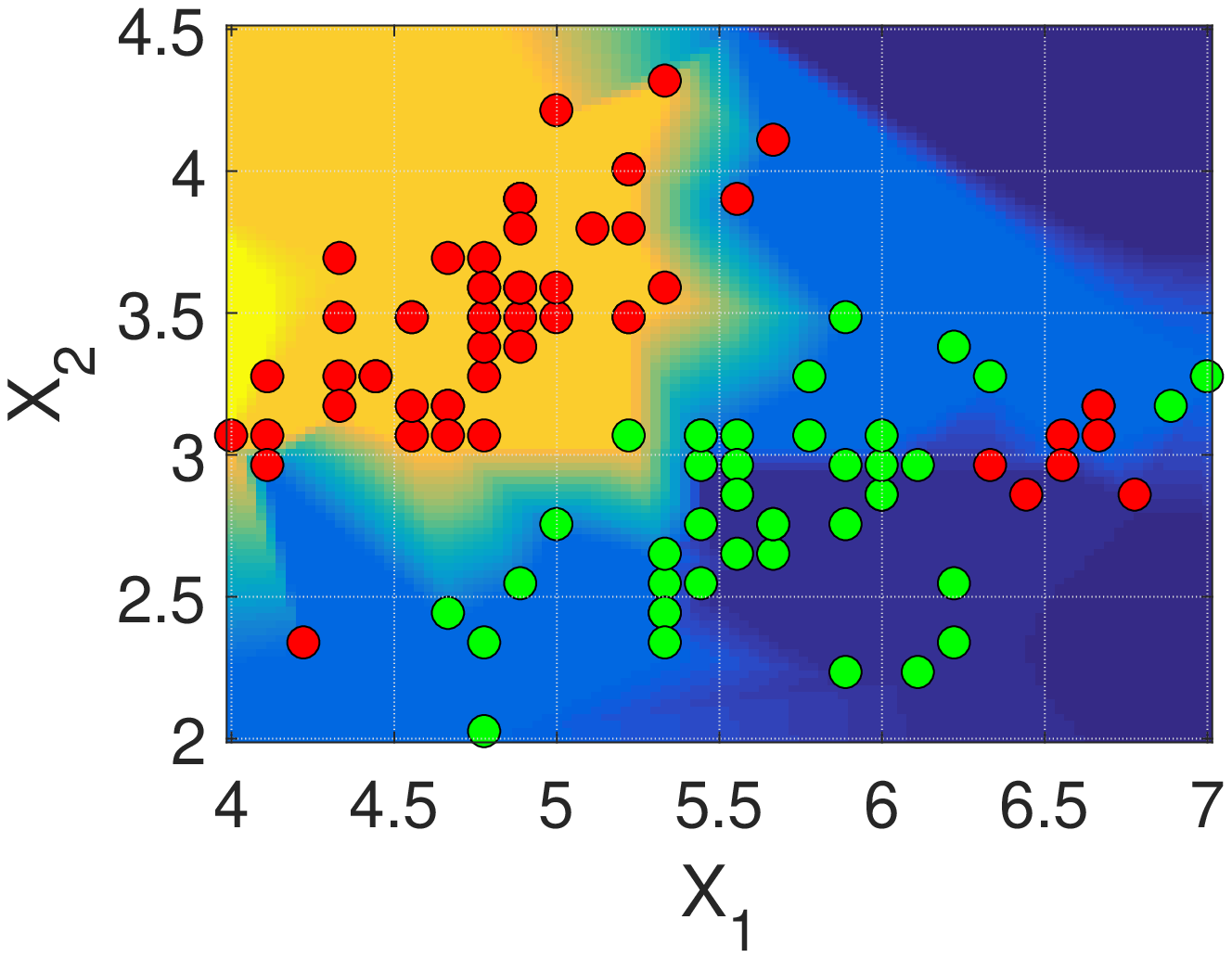}} 
\subfloat[$\lambda = 2$]{\includegraphics[width = 1.36in]{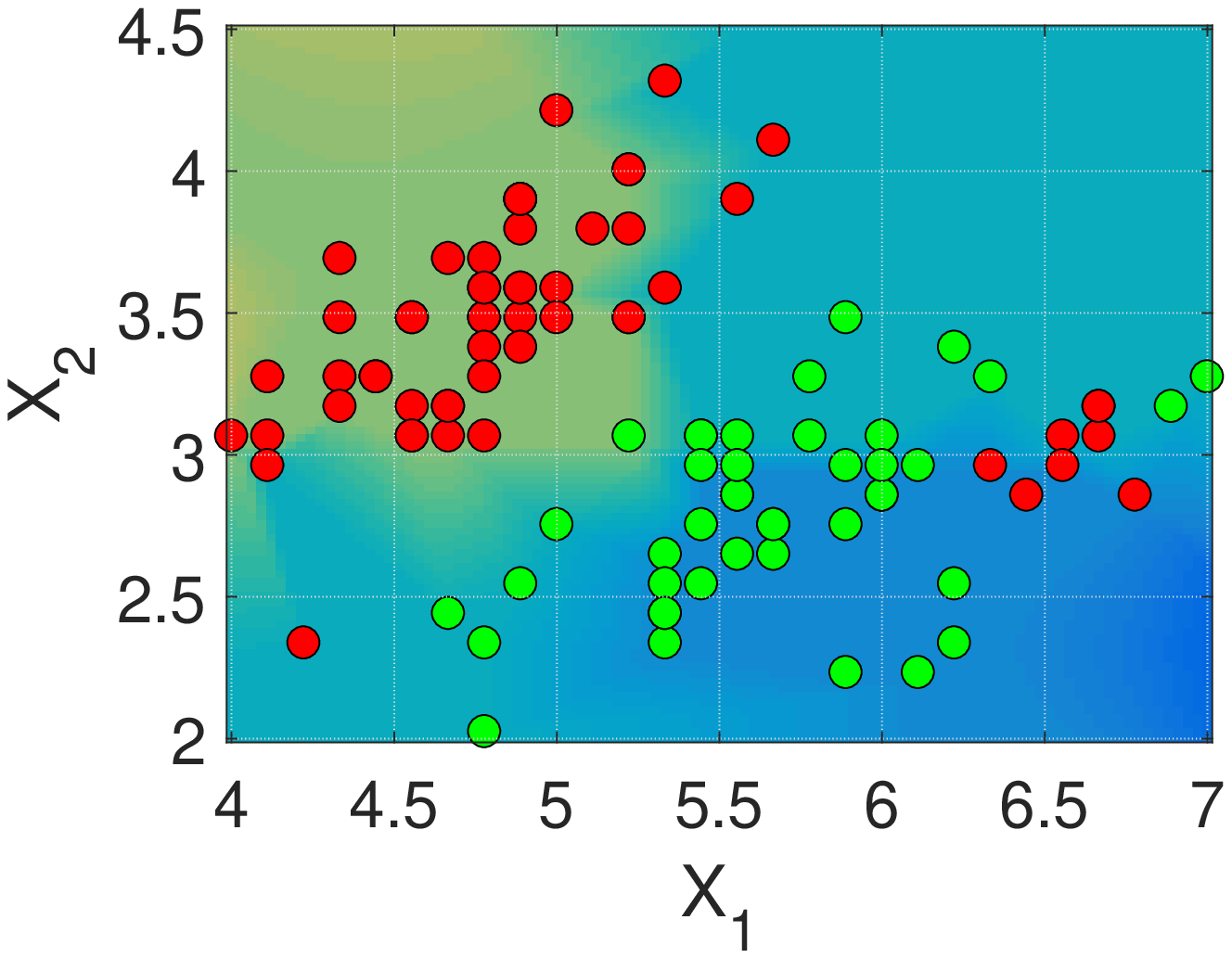}}
\subfloat[$\lambda = 20$]{\includegraphics[width = 1.36in]{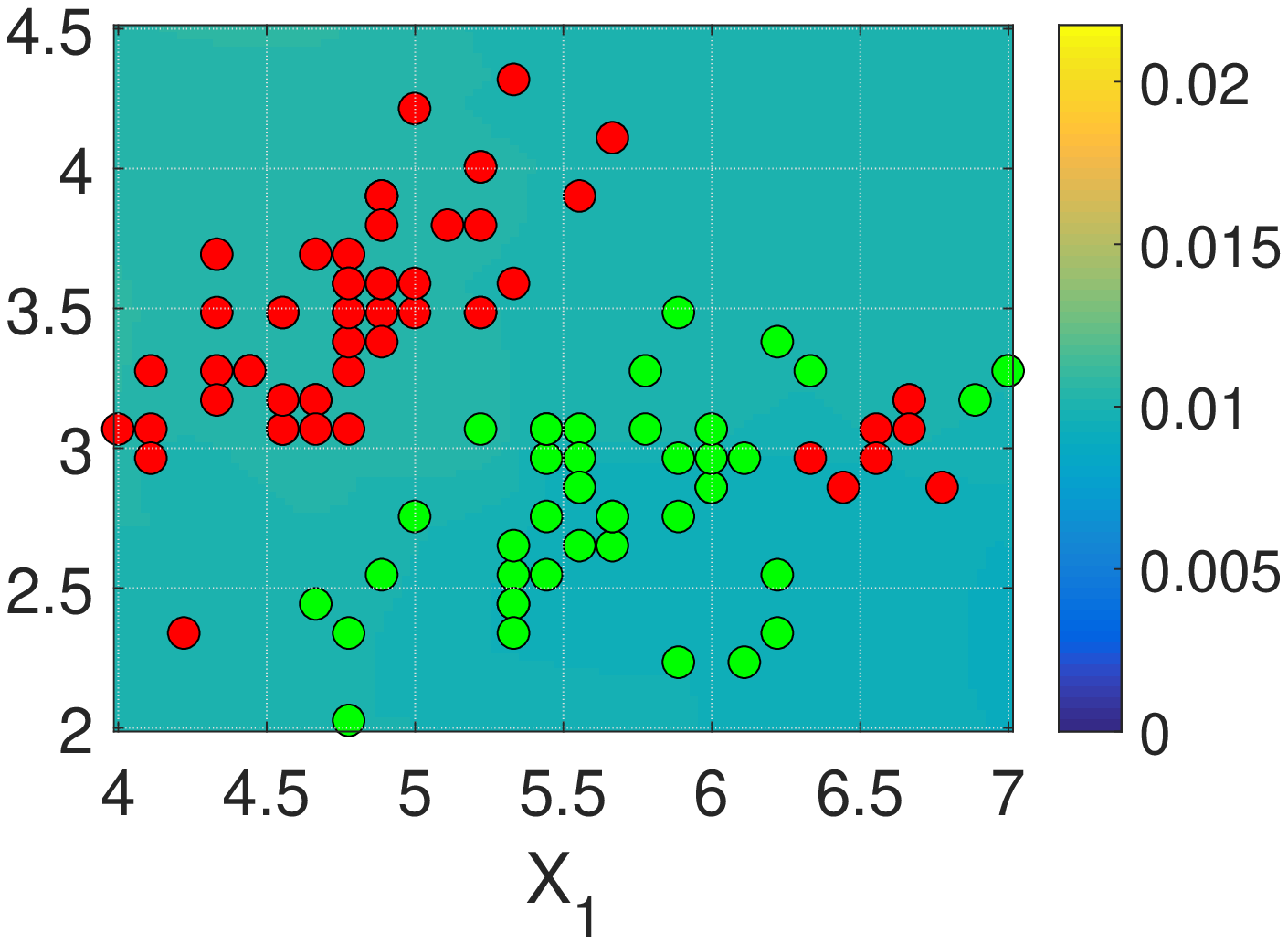}}
\caption{Heatmap of the instance weights for one leaf partition (upper left corner of each plot: $X_{2}>2.95 \wedge X_{1}<5.55$) at each of the instances (red and green points) as the value for $\lambda$ varies.}
\label{heat-weight}
\end{figure}

In a second experiment, we use a synthetic binary-labeled data set to graphically illustrate the  behavior of the instance weights as functions of lambda. The synthetic data set consists of 100 points in $\Reals^2$, out of which 58 belong to the red class, and the remaining 42 belong to the green class, as shown in Figure~\ref{heat-weight}.
The learning rate was chosen to be 0.1 based on classification accuracy, as in the previous experiment.  We recorded the instance weights produced by TSB at different values of $\lambda$.

Figure~\ref{heat-weight} shows a heatmap linearly interpolating the weights
associated with each instance for a disjoint region defined by one of the four leaf nodes of the trained tree. The
chosen leaf node corresponds to the logical function $(X_{2}>2.95) \wedge (X_{1}<5.55)$.

When $\lambda = 0$, the weights have binary normalized values that produce a sharp differentiation of the surface defined
by the leaf node, similar to the behavior of CART, as illustrated in Figure~\ref{heat-weight}(a). As $\lambda$ increases in value, the weights become more diffuse in Figures \ref{heat-weight} (b) and (c), until $\lambda$ becomes significantly greater than $1$. At that point,
the weights approximate the initial values as anticipated by theory.  Consequently, the ensembles along each path to a leaf are trained using equivalent instance weights, and therefore are the same and equivalent to GBS.

\section{Conclusions}

We have shown that tree-structured boosting reveals the intrinsic connections between additive models (GBS) and full interaction models (CART).  As the parameter $\lambda$ varies from $0$ to $\infty$, the models produced by TSB vary between CART and GBS, respectively.  This has been shown both theoretically and empirically.  Notably, the experiments revealed that a hybrid model between these two extremes of CART and GBS can outperform either of these alone.


\bibliographystyle{abbrvnat}
\bibliography{references}

\begin{thebibliography}{7}
\providecommand{\natexlab}[1]{#1}
\providecommand{\url}[1]{\texttt{#1}}
\expandafter\ifx\csname urlstyle\endcsname\relax
  \providecommand{\doi}[1]{doi: #1}\else
  \providecommand{\doi}{doi: \begingroup \urlstyle{rm}\Url}\fi

\bibitem[Breiman et~al.(1984)Breiman, Friedman, Olshen, and
  Stone]{breiman1984classification}
L.~Breiman, J.~H. Friedman, R.~A. Olshen, and C.~J. Stone.
\newblock \emph{Classification and regression trees}.
\newblock Wadsworth \& Brooks, Monterey, CA, 1984.

\bibitem[Caruana and Niculescu-Mizil(2006)]{Caruana2006Empirical}
R.~Caruana and A.~Niculescu-Mizil.
\newblock An empirical comparison of supervised learning algorithms.
\newblock In \emph{Intl. Conf. on Mach. Learn.}, pages 161--168, Pittsburgh,
  2006.

\bibitem[Friedman et~al.(1998)Friedman, Hastie, and
  Tibshirani]{Friedman1998Additive}
J.~Friedman, T.~Hastie, and R.~Tibshirani.
\newblock Additive logistic regression: a statistical view of boosting.
\newblock \emph{Annals of Stat.}, 28\penalty0 (2):\penalty0 2000, 1998.

\bibitem[Friedman(2001)]{friedman2001greedy}
J.~H. Friedman.
\newblock Greedy function approximation: a gradient boosting machine.
\newblock \emph{Annals of Stat.}, 29\penalty0 (5):\penalty0 1189--1232, 2001.

\bibitem[Lichman(2013)]{UCIMLRepository}
M.~Lichman.
\newblock {UCI} machine learning repository, 2013.
\newblock URL \url{http://archive.ics.uci.edu/ml}.

\bibitem[Loh(2014)]{Loh2014Fifty}
W.-Y. Loh.
\newblock Fifty years of classification and regression trees.
\newblock \emph{Intl. Stat. Review}, 82\penalty0 (3):\penalty0 329--348, 2014.

\bibitem[Valdes et~al.(2016)Valdes, Luna, Eaton, Simone~II, Ungar, and
  Solberg]{valdes2016mediboost}
G.~Valdes, J.~M. Luna, E.~Eaton, C.~B. Simone~II, L.~H. Ungar, and T.~D.
  Solberg.
\newblock {MediBoost}: a patient stratification tool for interpretable decision
  making in the era of precision medicine.
\newblock \emph{Scientific Reports}, 6, 2016.

\end{thebibliography}

\newpage

\appendix

\begin{center}
\large Supplemental Material for \\
\Large Tree-Structured Boosting: Connections Between Gradient Boosted Stumps and Full Decision Trees
\end{center}

\section{Proof Sketches of Key Lemmas}

This section shows that TSB is  equivalent to CART when $\lambda = 0$ and  equivalent to GBS as $\lambda \rightarrow \infty$, thus establishing a continuum between CART and GBS.  We include proof sketches for the four lemmas used to prove our main result in Theorem~\ref{thrm:main}.

\setcounter{lemma}{0}
\setcounter{equation}{0}

TSB maintains a perfect binary tree of depth $n$, with $2^{n}-1$ internal nodes, each of which corresponds to a weak learner. Each weak learner $h_k$ along the path from the root node to a leaf prediction node~$l$ induces two disjoint partitions of $\mathcal{X}$, namely $P_k$ and $P_k^c = \mathcal{X} \setminus P_k$ so that $ h_k(x_i) \neq h_k(x_j)$  $\forall$ $x_i \in P_k$ and $x_j \in P_k^c$. Let $\{R_{1}, \ldots, R_{n}\}$ be the corresponding set of partitions along that path to $l$, where each $R_{k}$ is either $P_k$ or $P_k^c$.
We can then define the partition of $\mathcal{X}$ associated with $l$ as $\bm{\mathcal{R}}_l = \setintersection_{k=1}^{n} R_{k}$.  TSB predicts a label for each $\vecx \in \bm{\mathcal{R}}_l$ via the ensemble consisting of all weak learners along the path to $l$ so that  $F(\vecx \in \bm{\mathcal{R}}_l) = \sum_{k=1}^{n} \rho_k h_k(\vecx)$. To focus each branch of the tree on corresponding instances, thereby constructing diverse ensembles, TSB maintains a set of weights $\bm{w} \in \Reals^N$ over all training data.  Let $\bm{w}_{n,l}$ denote the weights associated with training a weak learner $h_{n,l}$ at the leaf node $l$ at depth $n$. Notice that same as CART, TSB learns $2^n$ models, however, with better accuracy given its proven connection with GBS. 

We train the tree as follows. At each boosting step we have a current estimate of the function $F_{n-1}(\vecx)$ corresponding to a perfect binary tree of height $n-1$.
We seek to improve this estimate by replacing each of the $2^{n-1}$ leaf prediction nodes with additional weak learners $\{h'_{n,l}\}_{l=1}^{2^{n-1}}$ with corresponding weights $\bm{\rho}_n \in \Reals^{2^{n-1}}$, growing the tree by one level.  This yields a revised estimate of the function at each terminal node as
\begin{align}
\bm{F}_{n}(\vecx) = \bm{F}_{n-1}(\vecx) + \sum_{l=1}^{2^{n-1}} \rho_{n,l} \binary[\vecx \in \bm{\mathcal{R}}_l]
h_{n,l}'(\vecx) \enspace ,
\label{eqn:CombinedF}
\end{align}
where $\binary[p]$ is a binary indicator function that is 1 if predicate $p$ is true, and 0 otherwise.
Since partitions $\{\bm{\mathcal{R}}_1, \ldots, \bm{\mathcal{R}}_{2^{n-1}}\}$ are disjoint, Equation~\eqref{eqn:CombinedF} is equivalent to $2^{n-1}$ separate functions
\begin{displaymath}
F_{n,l}(\vecx \in \bm{\mathcal{R}}_l) = F_{n-1,l}(\vecx) + \rho_{n,l} h_{n,l}'(\vecx) \enspace ,
\end{displaymath}
one for each leaf's corresponding ensemble.  The goal is to minimize the loss over the data

\begin{equation}
L_n(\bm{X},\bm{y}) = \sum_{l=1}^{2^{n-1}} \sum_{i=1}^N w_{n,l,i}
\ell\!\left(y_i, F_{n-1,l}(\vecxi) +   \rho_{n,l} \binary[\vecxi \in \bm{\mathcal{R}}_l]
h_{n,l}'(\vecxi)\right), \label{eqn:CombinedLoss}
\end{equation}

by choosing $\bm{\rho}_n$ and the $h_{n,l}'$'s at each leaf. Taking advantage again of the independence of the leaves, Equation~\eqref{eqn:CombinedLoss} is minimized by independently minimizing the inner summation for each $l \in \{1,\ldots,2^{n-1}\}$, \emph{i.e.,}
\begin{align}
L_{n,l}(\bm{X},\bm{y}) &= \sum_{i=1}^{N}  w_{n,l,i} \ \ell\!\left(y_i, F_{n-1}(\vecxi) + \rho_{n,l}  h_{n,l}'(\vecxi)\right) \enspace .
\label{eqn:SeparateLossLeaf}
\end{align}
Note that \eqref{eqn:SeparateLossLeaf} can be solved efficiently via gradient boosting \cite{friedman2001greedy} of each $L_{n,l}(\cdot)$ in a level-wise manner through the tree.

Next, we focus on deriving TSB where the weak learners are binary regression trees with least squares as the loss function $\ell(\cdot)$.   Following Friedman~\cite{friedman2001greedy}, we first estimate the negative unconstrained gradient at each data instance $\left\{ \tilde{y}_i = - \frac{\partial \ell(y_i, F_{n-1}(\vecxi))}{\partial F_{n-1}(\vecxi)}\right\}_{i=1}^N$, which are equivalent to the residuals (i.e., $\tilde{y}_i = y_i - F_{n-1}(\vecxi)$).  Then, we can determine the optimal parameters for $L_{n,l}(\cdot)$ by solving
\begin{align}
\arg \min_{\rho_{n,l},h_{n,l}'} \sum_{i=1}^N w_{n,l,i} \   \Bigl(\tilde{y}_i - \rho_{n,l} h_{n,l}'(\vecxi)\Bigr)^2 \enspace .
\label{eqn:GBobjective}
\end{align}
Gradient boosting solves Equation~\eqref{eqn:GBobjective} by first fitting $h_{n,l}'$ to the residuals $(\bm{X},\bm{\tilde{y}})$, then solving for the optimal $\rho_{n,l}$. For details on gradient boosting, see \cite{friedman2001greedy}.  Adapting TSB to the classification setting, for example using logistic regression base learners and negative binomial log-likelihood as the loss function $\ell(\cdot)$, follows directly from \cite{friedman2001greedy} by using the gradient boosting procedure for classification in place of regression.

If all instance weights $\bm{w}$ remain constant, this approach would build a perfect binary tree of height~$T$, where each path from the root to a leaf represents the {\em same} ensemble, and so would be exactly equivalent to gradient boosting of $(\bm{X},\bm{y})$.  To focus each branch of the tree on corresponding instances, thereby constructing diverse ensembles, the weights are updated separately for each of $h_{n,l}$'s two children: instances in the corresponding partition have their weight multiplied by a factor of $1+\lambda$ and instances outside the partition have their weights multiplied by a factor of $\lambda$, where $\lambda \in [0, \infty]$.  The update rule for the weight $\wni$ of $\vecxi$ for $R_{n,l} \in \{P_{n,l}, P^c_{n,l}\}$ (the two partitions induced by $h_{n,l}$) is given by
\begin{equation}
  \wni = \frac{w_{n-1,l}(\vecxi)}{z_{n}}\Bigl(\lambda + \binary[\vecxi \in R_{n,l}]\Bigr) \enspace ,
  \label{update}
 \end{equation}
where $z_{n} \in \Re$ normalizes $\bm{w}_{n,l}$ to be a distribution. 
The initial weights $\bm{w}_{0}$ are typically uniform.


\setcounter{lemma}{0}
\setcounter{equation}{5}


\begin{lemma}
\label{lemma:weightDistribution}
 The weight of $\vecxi$ at leaf $l \in \{1, \ldots, 2^{n}\}$ at the $n^{th}$ boosting iteration ($\forall n = 1,2,\ldots$) is given by

 \begin{align}
    \wni = \frac{\wzeroi \left(\lambda + 1\right)^{\sum_{k=1}^{n}\binary[\vecxi \in R_{k}]}
	    \lambda^{\sum_{k=1}^{n}\binary[\vecxi \in R^{c}_{k}]}}{\prod_{k=1}^{n}z_{k}}  \enspace,
	    \label{weightn}
 \end{align}
where  $\{R_1, \ldots, R_n\}$ is the sequence of partitions along the path from the root to $l$.

\begin{proof}[Proof Sketch:] This lemma can be shown by induction based on Equation~\eqref{update}. \qed
\end{proof}
\end{lemma}

\begin{lemma} \label{lemmalimits}
 Given the weight distribution formula (\ref{weightn}) of $\vecxi$ at leaf $l \in \{1, ..., 2^n\}$ at the $n^{th}$ boosting iteration, the following limits hold,
  \begin{align}
  \lim_{\lambda \rightarrow 0} \wni &= \frac{\wzeroi}{\sum_{\vecx_{j} \in \capr}w_{0}(\vecx_j)}\binary[\vecxi \in \capr] \enspace, \label{limzero}\\
  \lim_{\lambda \rightarrow \infty} \wni &= \wzeroi \enspace.\label{liminf}
 \end{align}
 where  $\capr = \setintersection_{k=1}^n R_k$ is the intersection of the partitions along the path from the root to $l$.

\begin{proof}[Proof Sketch:] Both parts follow directly by taking the corresponding limits of Lemma~\ref{lemma:weightDistribution}. \qed
\end{proof}
\end{lemma}

\begin{lemma}\label{lemmareg}
The optimal simple regressor $\hn^{*}(\vecx)$ that minimizes the loss function (\ref{eqn:SeparateLossLeaf}) at the $n^{th}$ iteration at node $l \in \{1, \ldots, 2^n\}$ is given by,
\begin{equation}
 \hn^{*}(\vecx) = \left\{\begin{array}{lr}
                     \frac{\sum_{i:\vecxi\in \rbn}\wni \left(\yi-\Fnn(\vecxi)\right)}
                     {\sum_{i:\vecxi\in \rbn}\wni} &
                     \mbox{if } \vecxi \in \rbn \\ \\
                     \frac{\sum_{i:\vecxi \in \rcbn}\wni(\yi-\Fnn(\vecxi))}{\sum_{i:\vecxi \in \rcbn}\wni} & \mbox{otherwise}
                    \end{array}\right.. \label{regressor}
\end{equation}

\begin{proof}[Proof Sketch:]
For a given region $\rbn \subset \feat$ at the $n^{th}$ boosting iteration, the simple regressor has the form
 \begin{equation}
 h_{n}(\vecx) = \left\{\begin{array}{lr}
                   h_{n_{1}} & \mbox{if } \vecx \in \rbn \\ \\
                   h_{n_{2}} & \mbox{otherwise}
                  \end{array}\right.,
                  \label{hdef}
\end{equation}
with constants $h_{n_{1}},h_{n_{2}}\in \Reals$.
We take the derivative of the loss function \eqref{eqn:SeparateLossLeaf} in each of the two regions $\rbn$ and $\rcbn$, and solve for where the derivative is equal to zero, obtaining \eqref{regressor}. \qed
\end{proof}

\end{lemma}


\begin{lemma} \label{lemmaFconst}
The TSB update rule is given by
$F_{n}(\vecx) = \Fnn(\vecx) + \hn(\vecx)$. If $\hn(\vecx)$ is defined as,
\begin{align*}
\hn(\vecx) = \frac{\sum_{i:\vecxi\in \capr}\wni(\yi-\Fnn(\vecxi))}{\sum_{i:\vecxi\in \capr}\wni} \enspace,
\end{align*}
with constant $F_{0}(\vecx) = \bar{y}_{0}$, then $F_{n}(\vecx) = \bar{y}_{n}$ is \underline{constant}, with
\begin{equation}
 \bar{y}_{n} = \frac{\sum_{i:\vecxi\in \capr}\wni\yi}{\sum_{i:\vecxi\in \capr}\wni} \enspace, \ \ \ \ \forall n=1,2,\ldots \enspace . \label{yn}
\end{equation}

\begin{proof}[Proof Sketch:]
The proof is by induction on $n$, building upon \eqref{hdef}. We can show that each $h_n(\vecxi)$ is constant and so $\bar{y}_n$ is constant, and therefore the lemma holds under the given update rule. \qed
\end{proof}

\end{lemma}

Building upon these four lemmas, our main theoretical result is presented in the following theorem, and explained in the subsequent two remarks:
\begin{theorem}\label{thrm:main}
Given the TSB optimal simple regressor \eqref{regressor} that minimizes the loss function \eqref{eqn:SeparateLossLeaf},
the following limits regarding the parameter $\lambda$ of the weight update rule (\ref{update}) are enforced:
  \begin{align}
  \lim_{\lambda \rightarrow 0} \hn^{*}(\vecx) & =  \frac{\sum_{i:\vecxi\in \capr}\wzeroi\yi}
  {\sum_{i:\vecxi\in \capr}\wzeroi} - \bar{y}_{n-1}\label{CART}, \\ \nonumber \\
   \lim_{\lambda \rightarrow \infty}\!\hn^{*}(\vecx) & =  \left\{\!\!\begin{array}{lr}
                     \frac{\sum_{\vecxi \in \rbn}\wzeroi(\yi-\Fnn(\vecxi))}{\sum_{\vecxi \in \rbn}\wzeroi} & \mbox{if } \vecxi \in \rbn \\ \\
                     \frac{\sum_{\vecxi \in \rcbn}\wzeroi(\yi-\Fnn(\vecxi))}{\sum_{\vecxi \in \rcbn}\wzeroi} & \mbox{otherwise}
                    \end{array}\right.\label{gradBoosting}.
 \end{align}
 where $\wzeroi$ is the initial weight for the $i$-th training sample.
  \begin{proof}
  The limit (\ref{CART}) follows from applying (\ref{limzero}) from Lemma \ref{lemmalimits} to (\ref{regressor}) from Lemma \ref{lemmareg}
  regarding the result $F_{n}(\vecx) = \bar{y}_{n}$ with $\bar{y}_{n}$ a constant defined by (\ref{yn}) in Lemma \ref{lemmaFconst}. Similarly, the limit (\ref{gradBoosting})
  follows from applying (\ref{liminf}) from Lemma \ref{lemmalimits} to (\ref{regressor}) in Lemma \ref{lemmareg}.
  \qed
 \end{proof}
\end{theorem}

\begin{remark}
 \label{remark:CARTequivalency}
 The simple regressor given by (\ref{CART}) calculates a weighted average of the difference between the random output variables $y_{i}$
 and the previous estimate $\bar{y}_{n-1}$ of the function $F^{*}(\vecx)$ in the disjoint regions defined by $\capr$. This formally
 defines the behavior of the CART algorithm.
\end{remark}

\begin{remark}
\label{remark:GBequivalency}
 The simple regressor given by (\ref{gradBoosting}) calculates a weighted average of the difference between the random output
 variables $y_{i}$ and the previous estimate of the function $F^{*}(\vecx)$ given
 by the piece-wise constant function $\Fnn(\vecxi)$. $\Fnn(\vecxi)$ is defined in the overlapping region determined by
 the latest stump, namely $R_n$. This formally defines the behavior of the  GBS algorithm.
\end{remark}

\begin{remark}
\label{remark:MSEResEquivalency}
Without loss of generality, let us replace the loss function $\ell$ in \eqref{eqn:SeparateLossLeaf} by the squared error $\left( y_i - F_{n-1} \right)^{2}$ and set $\rho=1$. By replacing the functions $\hn^{*}(\vecx)$ given in Theorem \ref{thrm:main} at the limits of $\lambda$, the MSE (for $\lambda = 0$) and the residual (for $\lambda=\infty$) are enforced, which corresponds to the loss functions for CART and GBS respectively.
\end{remark}

\end{document}